\documentclass{article}
\usepackage{times}
\usepackage{epsfig}
\usepackage{graphicx}
\usepackage{amsmath}
\usepackage{amssymb}
\usepackage{booktabs}
\usepackage{textcomp}
\usepackage{subcaption}
\usepackage{soul}
\usepackage{algorithm}
\usepackage{algpseudocode}
\usepackage[table]{xcolor}
\usepackage{multirow}
\usepackage{tabularx}
\usepackage{tablefootnote}
\usepackage{url}
\usepackage[colorlinks]{hyperref}

\usepackage[inline]{enumitem}
\usepackage{natbib}

\newcommand{\ie}{\textit{i.e.}, }
\newcommand{\eg}{\textit{e.g.}, }

\newcommand{\starnote}[1]{}
\newcommand{\laststarnote}[1]{}
\newcommand{\donestarnote}[1]{}

\makeatletter
\DeclareRobustCommand{\change}{%
  \@bsphack
  \normalcolor %%% <<---- uncomment this line and comment the following two lines to go back to original black color
  \@esphack
}

\DeclareRobustCommand{\jchange}{%
  \@bsphack
  \normalcolor %%% <<---- uncomment this line and comment the following two lines to go back to original black color
  \@esphack
}

\DeclareRobustCommand{\stopchange}{%
  \@bsphack
  \normalcolor
  \@esphack
}
\makeatother

\title{Robotic Classification of Divers' Swimming States 
	using Visual Pose Keypoints as IMUs}

\author{
	Demetrious T. Kutzke\thanks{Science, Mathematics, and Research for Transformation (SMART)
		Scholarship provided through the US Department of Defense.} \\
	\textit{Department of Computer Science \& Engineering} \\
	\textit{Minnesota Robotics Institute (MnRI)} \\
	\textit{University of Minnesota--Twin Cities} \\
	\texttt{kutzk015@umn.edu}
	\and
	Ying-Kun Wu \\
	\textit{Department of Computer Science \& Engineering} \\
	\textit{Minnesota Robotics Institute (MnRI)} \\
	\textit{University of Minnesota--Twin Cities} \\
	\texttt{wu001210@umn.edu}
	\and
	Elizabeth Terveen \\
	\textit{Department of Computer Science} \\
	\textit{Carnegie Mellon University} \\
	\texttt{terve013@umn.edu}
	\and
	Junaed Sattar \\
	\textit{Department of Computer Science \& Engineering} \\
	\textit{Minnesota Robotics Institute (MnRI)} \\
	\textit{University of Minnesota--Twin Cities} \\
	\texttt{junaed@umn.edu}
}

\date{September 15, 2025}

\begin{document}

\maketitle
\thispagestyle{empty}
\pagestyle{empty}

\begin{abstract}
% Junaed's version below.
Traditional human activity recognition uses either direct image analysis or data from wearable inertial measurement units (IMUs), but can be ineffective in challenging underwater environments.
We introduce a novel hybrid approach that bridges this gap to monitor scuba diver safety. 
Our method leverages computer vision to generate high-fidelity motion data, effectively creating a ``pseudo-IMU'' from a stream of 3D human joint keypoints. 
This technique circumvents the critical problem of wireless signal attenuation in water, which plagues conventional diver-worn sensors communicating with an Autonomous Underwater Vehicle (AUV). 
We apply this system to the vital task of identifying anomalous scuba diver behavior that signals the onset of a medical emergency such as cardiac arrest---a leading cause of scuba diving fatalities.
By integrating our classifier onboard an AUV and conducting experiments with simulated distress scenarios, we demonstrate the utility and effectiveness of our method for advancing robotic monitoring and diver safety.

% Junaed's version ends here. 

% Visual human activity recognition often relies on deep learning methods directly in image space. Alternatively, smart devices or wearables use accelerometers and gyroscopes, or inertial measurement units (IMUs), for capturing human-body motion. Deep neural networks that exploit time series data perform inference directly on IMU data to classify human activities. In this work, we explore a hybrid approach that utilizes human joint keypoints in image space, \change treating the resulting temporal three-dimensional position estimates \stopchange as pseudo-IMUs for inference. \change This is analogous to the data obtained from diver-worn IMUs, which cannot transmit data to an autonomous underwater vehicle (AUV) because of the attenuation of wireless signals underwater. \stopchange We apply this approach for robotic classification of \change simulated ``anomalous'' human scuba diver behavior that could \stopchange indicate the onset of life threatening health conditions, such as cardiac arrest---a leading cause of scuba diving fatalities. We demonstrate the effectiveness of our method by integrating our system onboard an AUV and conducting semi-controlled experiments in closed-water environments. Results demonstrate the utility of robotic monitoring of scuba diver health during diving operations using our hybrid visual and time series classification method. 
\end{abstract}
% Accurate human diver pose estimation is fundamental for complex human-robot interaction scenarios, where the robot utilizes cues from human joint estimates to inform how the robot should move and interact with respect to the human. A robot utilizing a visual servo routine relies on consistent and robust pose estimation results for human joint localization to ensure smooth camera velocities to navigate from the robot's starting position to a final setpoint. Yet the underwater environment creates significant challenges for producing a smooth pose estimation result for every subsequent image frame from a robot's cameras. Typically, a pose keypoint will be perturbed within a pixel radius proportional to the error of the pose estimation network. We explore the extent to which statistical filtering and dynamics-based filtering can be used to smooth keypoint trajectories and provide keypoint interpolation when the pose estimation network fails to localize a human joint. This work ensures smooth, robust, and reliable pose estimation results without sacrificing processing speed. We perform in-lab experiments, measuring performance improvement of pose estimation networks using four different filtering techniques, and we perform in-water experiments using the best off-the-shelf pose estimation network to evaluate the efficacy of the robot's ability to smoothly move toward an ideal setpoint with respect to its human companion.
\section{Introduction}
\label{sec:introduction}
Detecting the onset of a life-threatening medical emergency in a scuba diver, such as cardiac arrest, is an immense challenge. 
Underwater, the clear physical symptoms that are obvious on land become obscured by the environment and equipment. 
While a companion autonomous underwater vehicle (co-AUV) acting as a robotic ``dive buddy'' is an ideal platform for continuous monitoring~\citep{islam2019understanding,codd2023recognizing,birk2022,demarco2014underwater}, communication remains a critical bottleneck. 
Conventional wireless technologies like WiFi and Bluetooth are highly attenuated in water, making it impossible to stream real-time health data from diver-worn sensors to the AUV.
Additionally, lower-frequency acoustic methods simply lack the bandwidth required for immediate inference~\citep{lundell2020diving,bosco2018environmental,dimmock2009risking}.

This work subverts that fundamental limitation of communicating diver health states.
Instead of relying on a data link from the diver, we enable the AUV to use its own vision to detect the most critical sign of severe medical events or disabling injuries (hereafter referred to as \textit{DI}) underwater: \textit{a sudden and complete cessation of movement}. 
An abrupt transition from swimming to a static, motionless state is a primary indicator that a diver is in distress and requires immediate intervention \citep{wilmshurst2020,Dipaolo2024}.

\begin{figure}
    \centering
    \includegraphics[width=1.0\columnwidth]{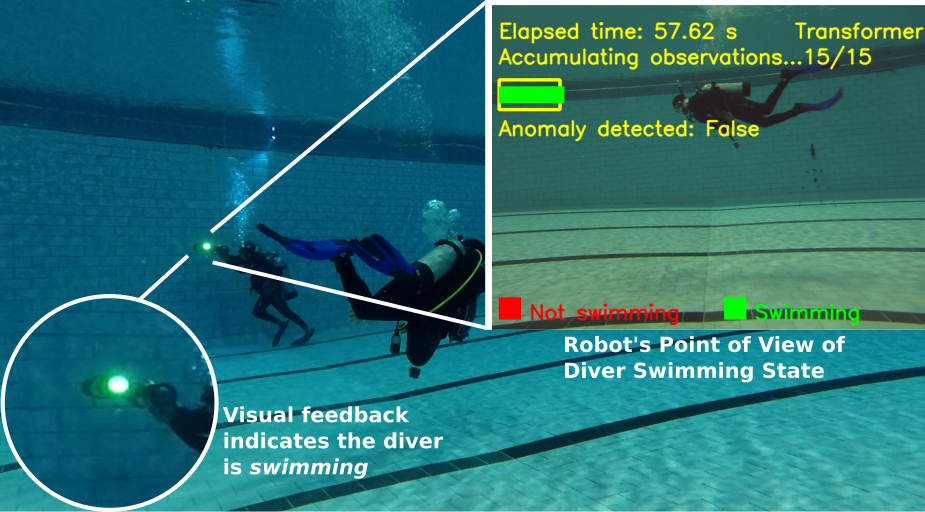}
        \caption{Temporal classification of scuba diver swimming state conducted during a closed-water evaluation of the diver anomaly classification system. The system is deployed on an AUV. The AUV provides visual feedback of predicted state by illuminating a series of concentric LED lights controlled to reflect a green color which indicates the diver is swimming. }
        \label{fig:cover_image}
    \vspace{-6mm}
\end{figure}

To achieve this, we introduce a novel method that transforms the AUV's visual feed into a virtual motion sensor. 
Fig.~\ref{fig:cover_image} shows the proposed system during an in-water evaluation of a diver's swimming state.
We use three-dimensional human pose estimations from a monocular camera to approximate \textit{translational} acceleration of the diver's joint keypoints.
Additionally, we establish a body frame convention to estimate the diver's \textit{rotational} acceleration using the same body keypoints. 
Together, these constitute a uniquely novel approach of using human pose joints for scuba diver swimming state, and eventually health state, classification. 
By tracking the diver's 3D joint keypoints over time, we generate a stream of translational and rotational acceleration data--effectively creating a ``pseudo-inertial measurement unit'' (pseudo-IMU) from vision alone. 
This allows a deep learning model to classify the diver's swimming state in real-time without any instrumentation on the diver.

In this paper, we make two primary contributions: (1) a novel system for classifying a scuba diver's swimming state using only a monocular camera on an AUV, and (2) a unique and diverse dataset of underwater human motion, capturing the transition from swimming to non-swimming states. 
We demonstrate the effectiveness of our system through in-water experiments, proving it is a viable and powerful new approach for robotic monitoring of diver safety.
\section{Literature Review}
\label{sec:lit_review}

Our work integrates two primary domains: human pose estimation and time-series classification.

\textbf{Human Pose Estimation}. Human pose estimation is the process of localizing human joints in images, typically using Deep Neural Networks (DNNs) with convolutional layers to model the topological relationships between limbs~\citep{zheng2023deep}. 
For this work, we require 3D keypoint data to estimate accelerations. 
3D pose estimation networks often generate these coordinates by applying triangulation to 2D poses or by training on data with labeled depth information~\citep{lee2018propagating, nunez2019multiview, wang2021deep}. 
Performance can be improved by leveraging temporal data from previous frames or through modern attention-based models~\citep{zheng20213d}.

\begin{figure}[h]
\vspace{1.75mm}
\captionsetup[subfigure]{labelformat=empty}
    \centering
    % \vspace{2mm}
    \begin{subfigure}[t]{0.48\columnwidth}
        \includegraphics[width=1.0\textwidth]{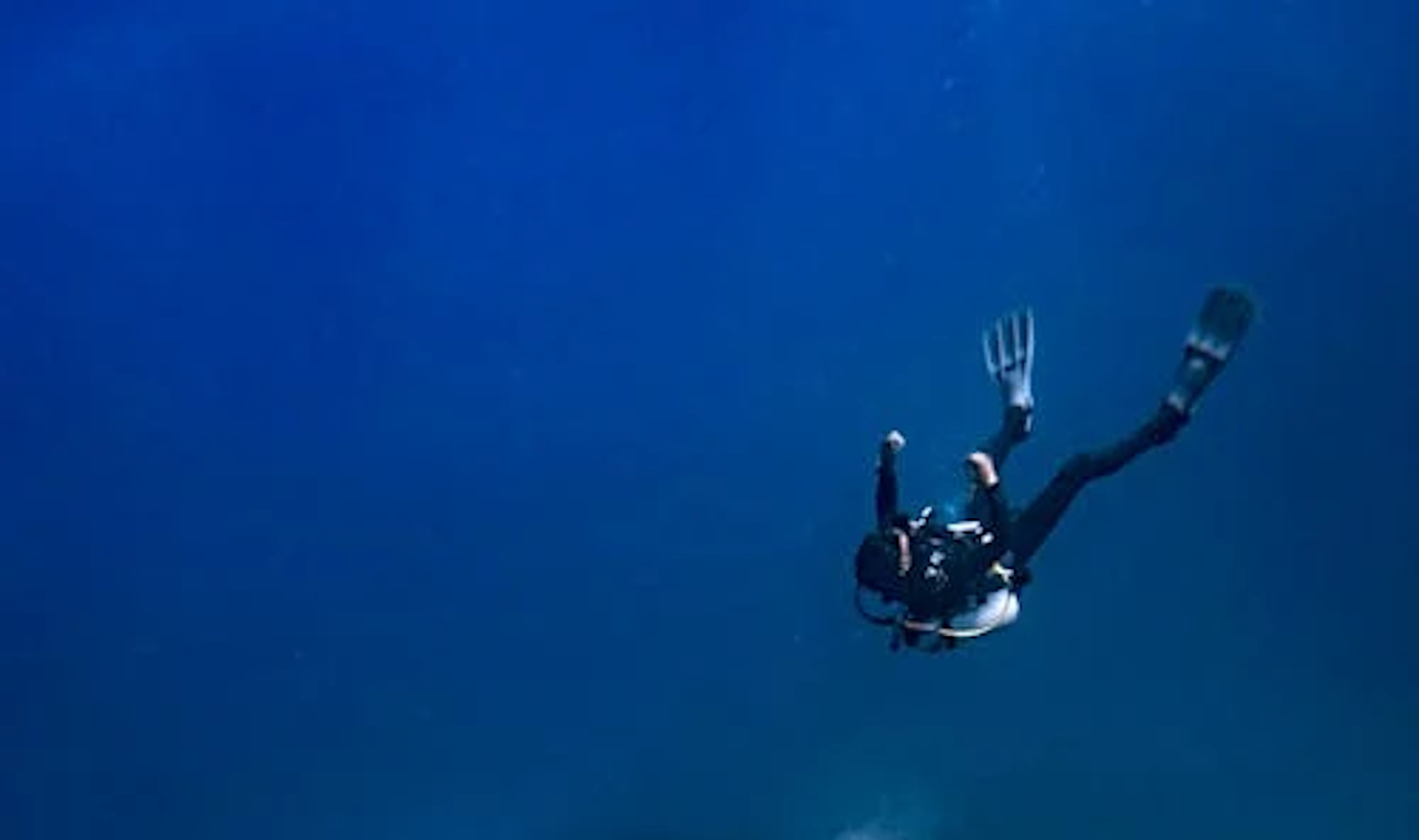}
        \caption{Prone up}
    \end{subfigure}
    \begin{subfigure}[t]{0.48\columnwidth}
        \includegraphics[width=1.0\textwidth]{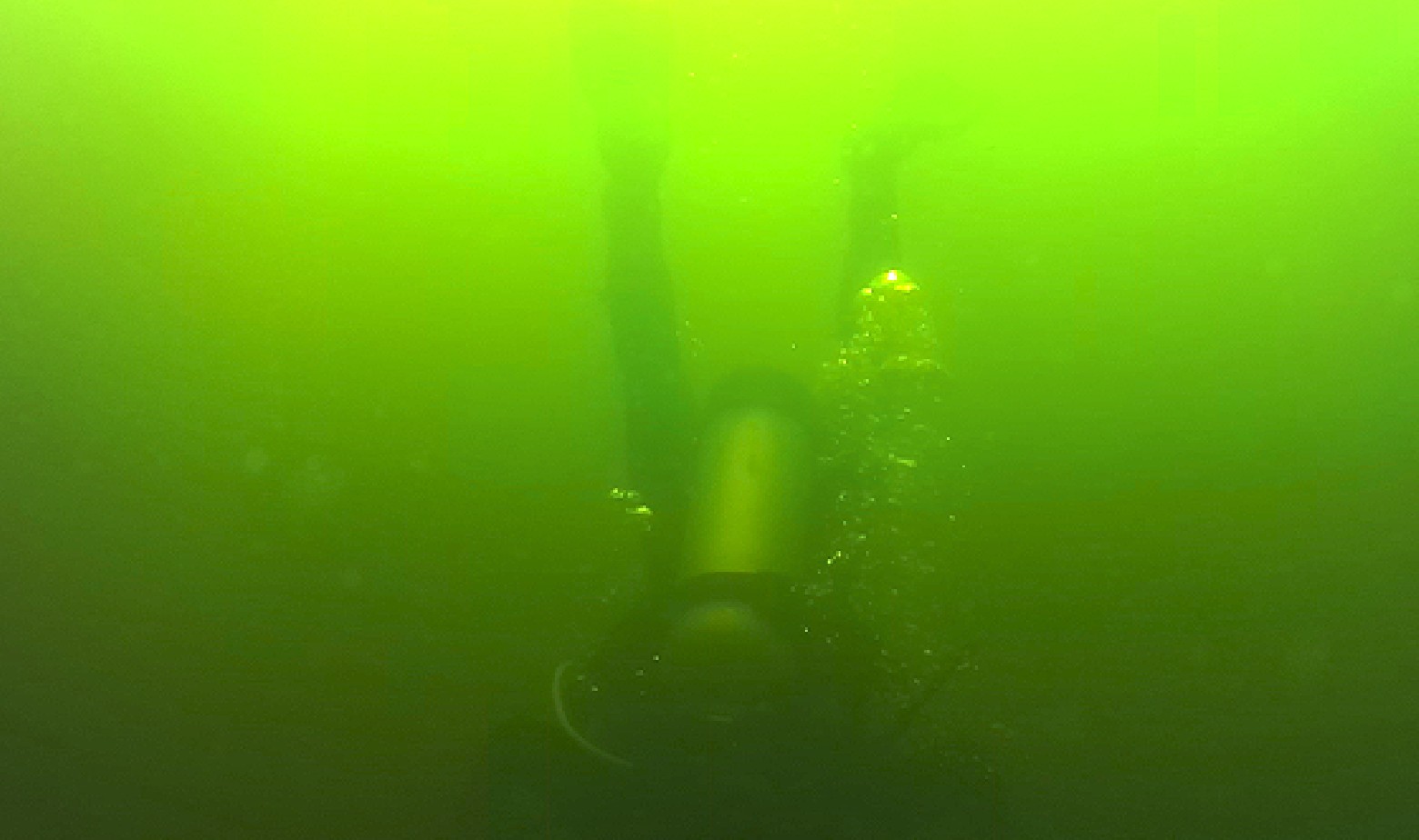}   
        \caption{Inverted}        
    \end{subfigure}  
    \begin{subfigure}[t]{0.48\columnwidth}
        \includegraphics[width=1.0\textwidth]{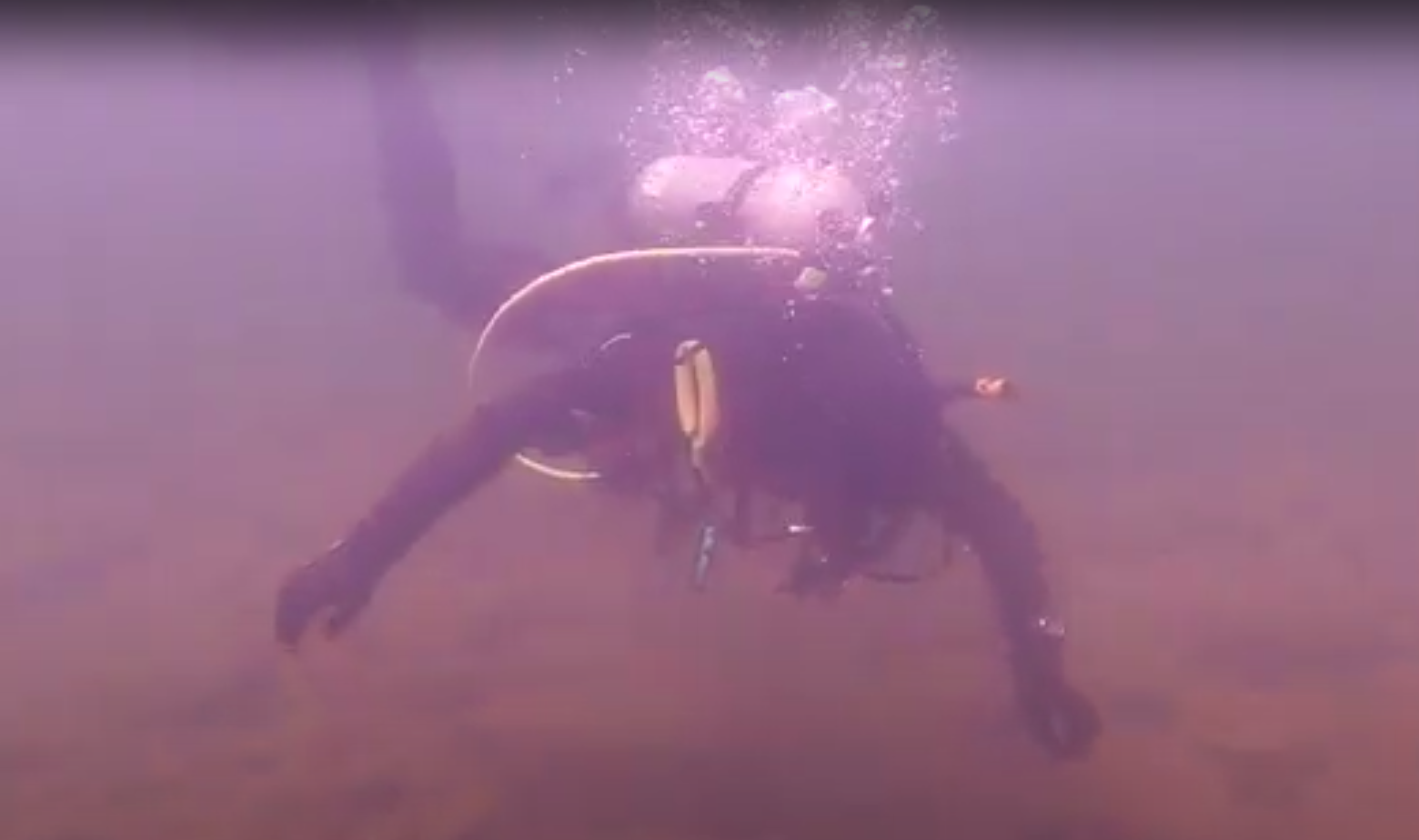}   
        \caption{Prone down}        
    \end{subfigure}
    \begin{subfigure}[t]{0.48\columnwidth}
        \includegraphics[width=1.0\textwidth]{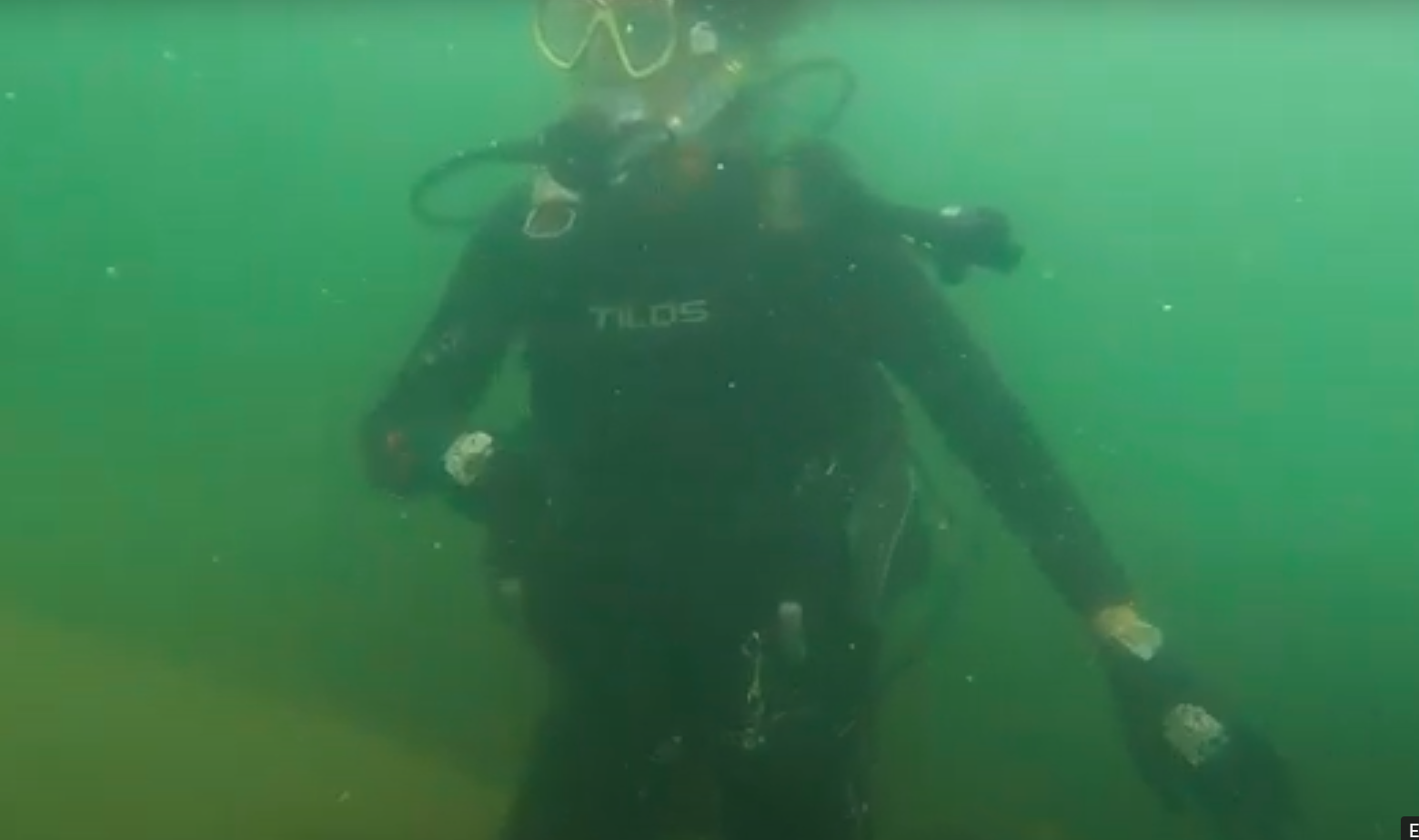}
        \caption{Upright}        
    \end{subfigure}
    \caption{Typical diver poses during scuba diving operations. Deep neural networks trained on terrestrial data often fail to localize human joints for these poses, given the uniqueness of the joint configurations and visibility.}
    \label{fig:nonstandard_pose}
\vspace{-3.5mm}
\end{figure}

The underwater environment is a unique challenge, since divers can adopt poses in six degrees of freedom (6DOF) that are highly non-standard in the terrestrial domain (see Fig.~\ref{fig:nonstandard_pose}). 
Off-the-shelf models trained on terrestrial data often fail to localize joints in these unique configurations. 
We overcame this by fine-tuning a \textit{YOLOv8} model~\citep{ultralyticsyolov8} on a hand-labeled underwater dataset and then using the \textit{VideoPose3D}~\citep{pavllo20193d} estimator, trained on AIST++~\cite{aist++} and H3WB~\citep{h3wb}, to lift the resulting 2D poses to 3D.
While a recent method has emerged for extracting 3D poses underwater using stereo vision without labeled data~\citep{wu2025stereo}, our approach has the distinct advantage of requiring only a monocular camera.

\textbf{Time-Series Classification}. Time series classification (TSC)~\citep{schmidl2022_anomaly,ismail2019deep} assigns a class label to a sequence of data. 
A related subproblem is time series anomaly detection (TSAD), which identifies anomalous points within a sequence. 
% There are two prevailing methods for labeling anomalous sequences.\starnote{@DTK: Could you clarify what two methods are we talking about? DT: no need for this sentence.}
A collective anomaly~\citep{bontemps2016collective} refers to the classification of a sequence as anomalous considering the aggregate effect of multiple underlying measurements, even though an individual measurement might not be considered anomalous.
Multiple methods exist for detecting collective anomalies, including \citet{keogh2005} and \citet{yeh2023sketching}, both of which utilize discords for anomalous subsequence detection in collective samples, effectively decreasing inference time by ignoring irrelevant dimensions. 
\change Anomaly detection is a difficult problem due to the paucity of data available on rare events. 
This is heightened in settings for which direct measurement of anomalous events is either challenging, impossible, or too rare to capture substantive data. 
% For example, while hospitals may have ample access to anomalous echocardiograms---a key diagnostic tool for cardiac events----given the sheer number administered annually \cite{ferraz2023_echocardiogram}, 
Collecting vital signs of scuba divers during diving operations is such an example, unless conducted under strict experimental settings and with specialized equipment \citep{keuski2018updates}\stopchange.

Though our goal is to detect health anomalies, our method employs TSC rather than TSAD. 
This allows us to classify an entire sequence as a ``swimming state'' without needing to precisely engineer the specific features that constitute an anomaly~\citep{garg2021evaluation}. 
By collecting data in structured, uniform intervals, we were able to create representative class sequences without manual segmentation, using pseudo-IMU values as swimming state. 
We followed the method described in~\citet{hu2016} to create subsequences of representative data manually, eliminating the need to do feature engineering by hand.

A central challenge in TSC is simultaneously capturing short-term and long-term dependencies. 
Hybrid models~\citep{foumani2024, fawaz2019} combining Convolutional and Recurrent Neural Networks (CNNs, RNNs) are effective at this, as they merge the strengths of each architecture while mitigating their individual drawbacks (\eg a CNN's failure to capture long-term patterns or an RNN's computational expense). 
Attention-based models also excel at capturing long-term dependencies~\citep{foumani2024}, though often at a higher computational cost, especially for embedded systems~\citep{2024edgeattention}.

\section{Data Collection and Feature Extraction}
\label{sec:data_collection}
Data collection consisted of two parts: (1) non-standard body pose data collection, which includes collection of images to perform camera calibration to extract camera intrinsic and extrinsic parameters, and  (2) diver swimming state transition data from swimming to not swimming. Data was collected in accordance with Institutional Review Board (IRB) regulations and assessed to be not human research.

% reference No. 0017807, 

% assessed to be not human research.

\textbf{A Note About the Ethics of In-Water Data Collection}. We cannot ask participants to stop moving in open water environments, since swimming, kicking, and paddling are critical for maintaining good buoyancy or depth control in the water column. Without buoyancy control, divers are at risk of significant medical issues such as barotrauma injuries from uncontrolled descents or decompression sickness due to gas embolisms from uncontrolled ascents \citep{carlston2012physician}. To ensure diver safety, we performed data collection and in-water evaluations of our robotic system in a closed-water swimming facility. This allowed us to mitigate risks to the diver as well as decouple the efficacy of the proposed system from environmental conditions. 
\jchange However, in Section~\ref{sec:open_water}, we present initial results from ongoing field tests, along with a discussion on planned future improvements, demonstrating the generalization capability of our method to challenging open-water environments.\stopchange
% \change We acknowledge that the dataset and results presented have limitations, given the pose and anomaly data were collected in a closed-water swimming facility. However, an ongoing effort is to show our methodology generalizes to open-water environments by conducting continued field experiments. Section~\ref{sec:open_water} presents initial results from field testing and a discussion on planned future improvements.\stopchange
% \vspace{4mm}

% \starnote{JS: Elizabeth has good points in overleaf comments, please see and address.}
\textbf{Non-standard Body Pose Data}. Divers can achieve highly atypical body poses underwater compared to the terrestrial domain (see Fig.~\ref{fig:nonstandard_pose}). To ensure that our pose estimation network could accommodate non-standard body poses, we collected and aggregated 3305 stereo pair images of resolution $640\times360$ pixels at $10$~fps; these include divers in four primary poses over different viewpoints: \textit{prone down}, \textit{prone up}, \textit{inverted}, and \textit{upright}.
Note that we use a stereo camera effectively only to double the size of the dataset; our method only uses single images for inference, and camera calibration is used only to remove geometric distortions. 
%Calibration is used only to remove geometric distortions. \stopchange %  from each stereo pair of images\stopchange

\begin{figure}[h]
    \centering
\includegraphics[width=1.0\columnwidth]{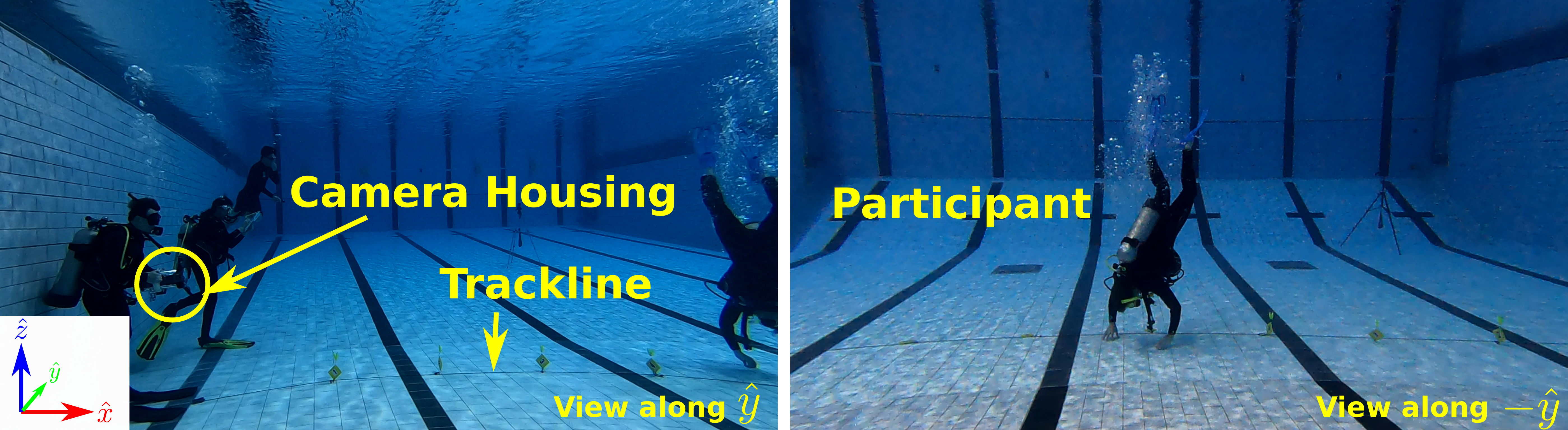}
    \caption{(Best if viewed at $3\times$ zoom level). Experimental setup for in-water data collection of non-standard body pose data and diver state transition data. The trackline delineates distance between the camera and participant. We collected image data at a depth of approximately $3.5$~m and between $3$ to $5$~m from the camera's image plane.}
    \label{fig:experimental_setup}
    \vspace{-5mm}
\end{figure}

% \change To be clear, we collected stereo camera data to expand our visual image dataset size. The method introduced in this paper is a monocular method that does not require a stereo camera setup. \stopchange
Each diver was asked to rotate $360$ degrees about the $\hat{z}$-axis in place while maintaining a given body pose. Divers were asked to do this at distances between $3$ and $5$ meters from the camera as measured by a trackline with visible markers extending from the camera image plane, along the camera frame's $\hat{x}$-axis. Fig.~\ref{fig:experimental_setup} demonstrates the data collection setup. 

\begin{figure}[h]
    \centering
    \includegraphics[width=0.85\columnwidth]{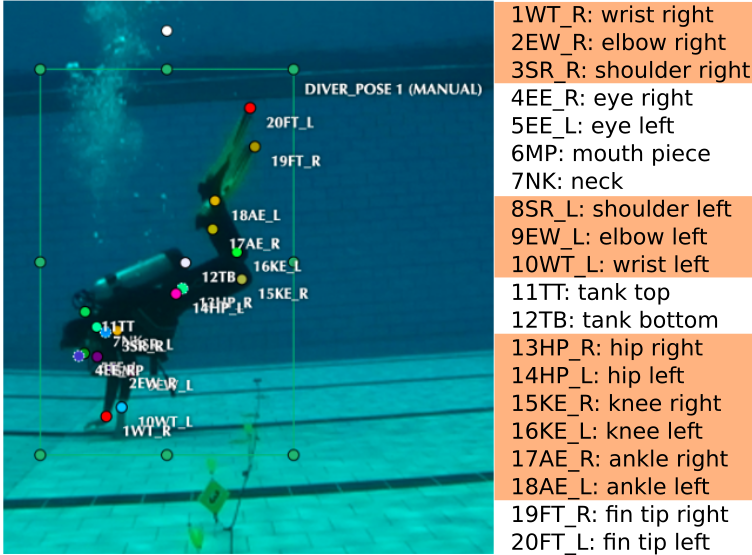}
    \caption{(Best if viewed at $1.5\times$ zoom level). Pose keypoints labeled using an augmented COCO convention. We utilize the keypoints highlighted in orange in our analysis.}
    \label{fig:pose_keypoints}
    \vspace{-2mm}
\end{figure}

We labeled the images utilizing an augmented COCO convention \citep{lin2014microsoft}. We labeled $20$ pose keypoints shown in Fig.~\ref{fig:pose_keypoints}. 
These keypoints include objects relevant to most recreational divers such as the mouthpiece, top of the tank and bottom, and fin tips. For our analysis, we selected the $12$ keypoints corresponding to the diver's major joints (shoulders, elbows, wrists, hips, knees, and ankles), highlighted in Fig.~\ref{fig:pose_keypoints} in orange. This set was chosen to capture the dynamics of the limbs and torso, which are the primary drivers of propulsive and stabilizing movements underwater, while excluding less informative points like the eyes or external equipment.

% For this work, we utilized $12$ relevant keypoints, excluding the eyes, neck, and dive related equipment, because the keypoints which fall on the diver's body are more representative of core body movement. We highlight the used keypoints in Fig.~\ref{fig:pose_keypoints} in orange.
% \vspace{4mm}

% An underwater apparatus made from acrylic encapsulates the camera (labeled as ``camera housing" in Fig. ~\ref{fig:experimental_setup}). The acrylic face, which is planar and parallel with the stereo camera pair, introduces distortion which couples to the inherent scattering, distortion, and other optical effects underwater \cite{mobley1995optical}.

\textit{YOLOv8} requires undistorted images for training and inference, and \textit{VideoPose3D} requires a sequence of calibrated two-dimensional poses. To ensure that we obtained accurate pose estimates for our method, we collected three sets of calibration data to compute camera intrinsic and extrinsic matrices. We utilized three different calibration target sizes, including $4\times6$, $5\times8$, and $8\times10$ tags in a grid, using the \textit{Kalibr} target specification \citep{oth2013rolling}. Specific target dimensions are included in the accompanying video. %(Fig.~\ref{fig:calibration})

% \begin{figure}[h]
% \vspace{1.75mm}
% \captionsetup[subfigure]{labelformat=empty}
%     \centering
%     % \vspace{2mm}
%     \begin{subfigure}[t]{0.32\columnwidth}
%         \includegraphics[width=1.0\textwidth]{Figures/PNG_Sources/zed_2024-06-16-19-15-33_4x6_example_image.png}
%         \caption{}
%     \end{subfigure}
%     \begin{subfigure}[t]{0.32\columnwidth}
%     \includegraphics[width=1.0\textwidth]{Figures/PNG_Sources/zed_2024-06-16-19-15-33_5x8_example_image.png}   
%         \caption{}        
%     \end{subfigure}  
%     \begin{subfigure}[t]{0.32\columnwidth}
%         \includegraphics[width=1.0\textwidth]{Figures/PNG_Sources/zed_2024-06-16-19-15-33_6x10_example_image.png}
%         \caption{}        
%     \end{subfigure}
%     \vspace{-5mm}
%     \caption{Camera calibration using the \textit{Kalibr} specification with three different calibration target sizes: $4\times6$, $5\times8$, and $6\times10$ tags, from left to right, respectively.}
%     \label{fig:calibration}
% \vspace{-4mm}
% \end{figure}

\begin{figure}[h]
\vspace{-2mm}
\captionsetup[subfigure]{labelformat=empty}
    \centering
    % \vspace{2mm}
    \begin{subfigure}[t]{0.48\columnwidth}
\includegraphics[width=1.0\textwidth]{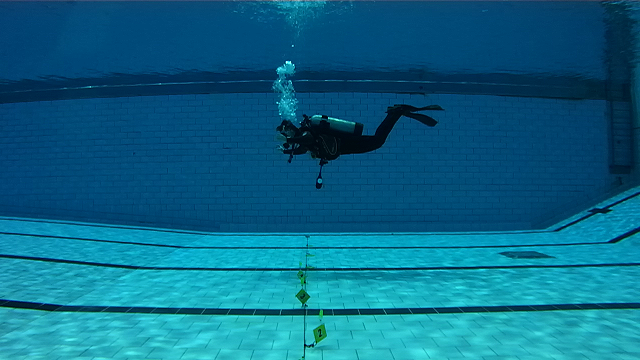}
        \caption{Swimming}
    \end{subfigure}
    \begin{subfigure}[t]{0.48\columnwidth}
    \includegraphics[width=1.0\textwidth]{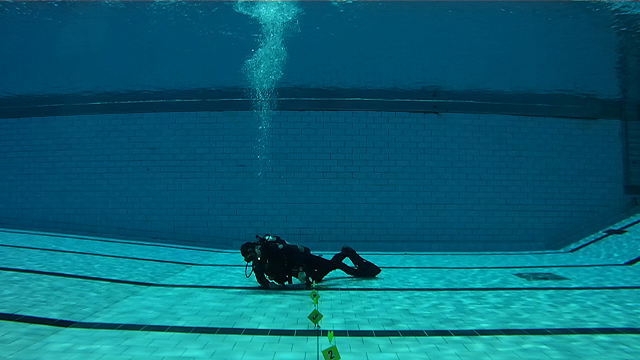}   
        \caption{Not swimming}        
    \end{subfigure}  
    \caption{Example moving and not moving images. In this case, the diver is hovering \textit{prone down}, and when they stop moving they sink.}
    \label{fig:moving_not_moving}
\vspace{-3mm}
\end{figure}

\textbf{Diver State Transition Data}. 
% Fig.~\ref{fig:data_collection_setup} demonstrates our experimental setup. First, three divers wer
Divers were asked to maintain each of the four primary poses (\textit{prone down}, \textit{prone up}, \textit{inverted}, and \textit{upright}) for approximately $10$~s by swimming, kicking, or paddling as necessary to maintain that pose. Then they were asked to stop these movements for $5$~s. This was done from different viewpoints around the $\hat{z}$-axis as in Fig.~\ref{fig:experimental_setup}. By stopping normal swimming activities, this closely mimics the phenomena of sudden cardiac arrest or other DI. Based on the buoyancy adjustment of the participant, sometimes the participant would sink, rise to the surface, or remain in place. Fig.~\ref{fig:moving_not_moving} shows one case in which the diver sank to the bottom during the no movement phase of data collection.

\textbf{Feature Extraction}. \jchange The cornerstone of our method is the conversion of raw 3D keypoint coordinates into a meaningful, motion-based feature vector. \stopchange  \change This process involves two key steps: first, estimating up-to-scale 3D keypoints from divers using off-the-shelf estimators, and second, establishing a stable body-centric reference frame to compute the diver's overall translational and rotational acceleration.\stopchange 

Let the input to the pose estimation network be an undistorted image of width $W$, height $H$, and resolution $W\times H$. Let $\mathbf{K}=\{\mathbf{k}_1,\dots, \mathbf{k}_{M}\}$ be the set of predicted human joint keypoints in image coordinates, \ie $\mathbf{k} = (u,v)\in[0,W-1]\times[0,H-1]$. Note that we utilize $M=12$ different keypoints defined in Fig.~\ref{fig:pose_keypoints}. \change The 2D-to-3D lifting network then transforms these predictions into 3D space with arbitrary scale. \stopchange Now let $\mathbf{R}=\{\mathbf{r}_1,\dots, \mathbf{r}_{M}\}$ be the set of corresponding human joint keypoints in three-dimensional coordinates with respect to the camera frame, \ie $\mathbf{r}=(x,y,z)\in \mathbb{R}^{3}$. One can show that using a Taylor series expansion, the \textit{translational acceleration} in the $x$-direction $\ddot{x}$ is approximately 
\begin{equation}
    \ddot x \approx \frac{x_{i+1}-2x_{i}+x_{i-1}}{\Delta t^{2}},  
\end{equation}
where $\Delta t$ is the temporal separation of image frames, which is effectively the inverse of the frame rate of the camera. $x_i$ is the measurement of the $x$ position at time step $i$. Similar formulas exist for the $\hat{y}$- and $\hat{z}$-directions. 

\begin{figure}[t]
\vspace{2mm}
\captionsetup[subfigure]{labelformat=empty}
    \centering
    % \vspace{2mm}
    \begin{subfigure}[t]{0.48\columnwidth}
    \centering
        \includegraphics[width=1.0\linewidth]{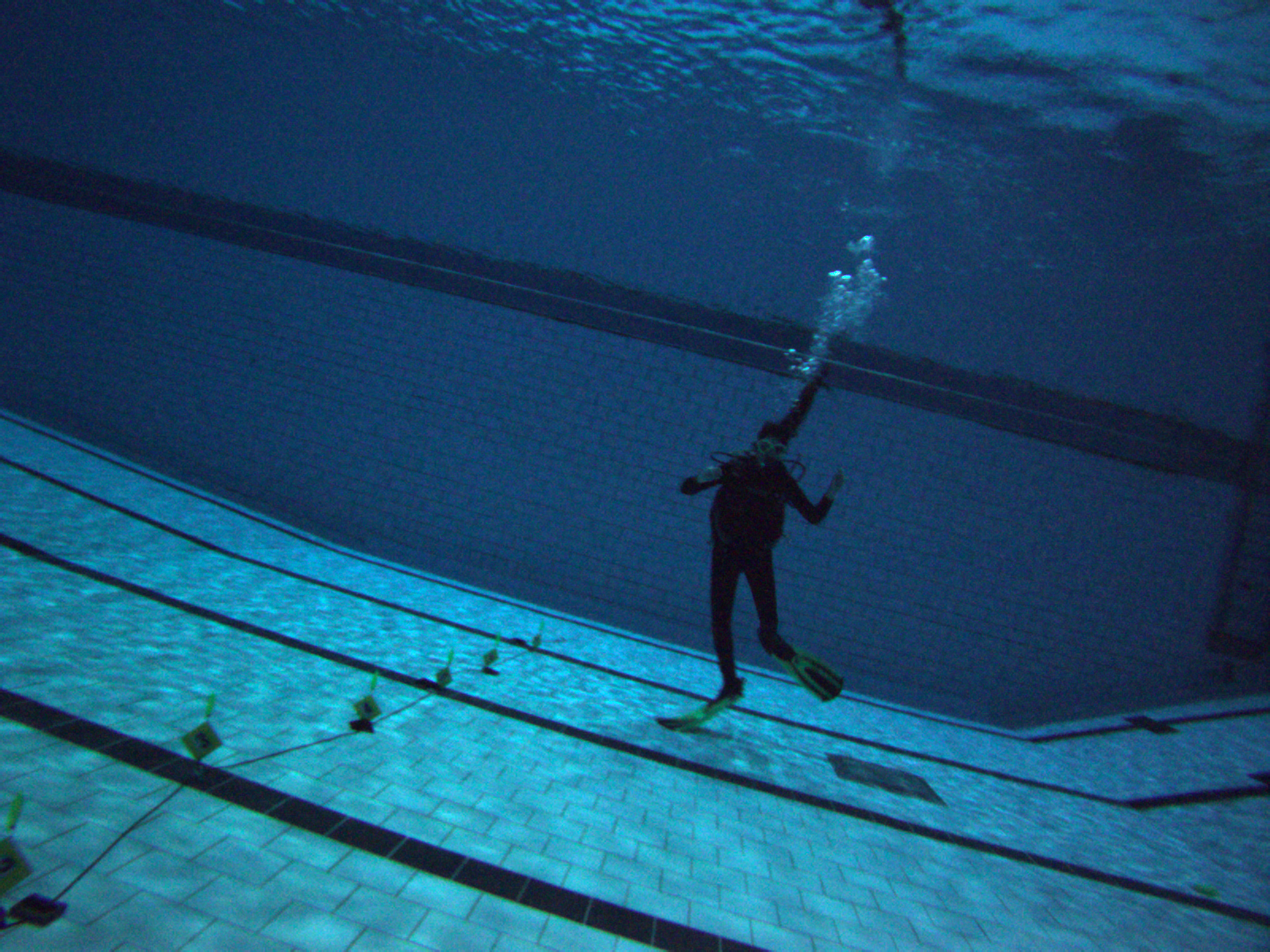}
        \caption{Raw image (brightened for visibility).}
    \end{subfigure}
    \begin{subfigure}[t]{0.48\columnwidth}
    \centering
    \includegraphics[width=1.0\linewidth]{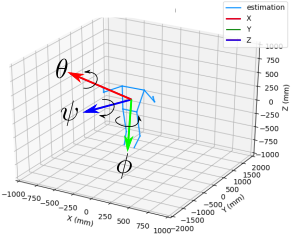}   
        \caption{Body frame showing the rotation angles.}
    \end{subfigure}  
    \caption{(Best if viewed at $2\times$ zoom level). Raw input image and the body frame with rotation angles after three-dimensional pose estimation.}
    \label{fig:body_frame}
\vspace{-5mm}
\end{figure}

%To compute the \textit{rotational acceleration} of the diver's body, we utilize a frame convention that is affixed to the diver's torso (see  Fig.~\ref{fig:body_frame}). 
\jchange To accurately measure the diver's self-initiated rotation, we must first decouple their movements from the motion of the AUV's camera. We achieve this by defining a dynamic body-centric reference frame affixed to the diver's torso (see  Fig.~\ref{fig:body_frame}). This ensures our rotational acceleration features are invariant to the robot's position and orientation.\stopchange 
The frame convention relies on the torso pose keypoints of the diver $\mathbf{r}_{\text{torso}} \in \mathbf{R}$ , creating a frame that is located at the approximate center of the human's chest. Let $\mathbf{r}_{\text{torso}} = \{\mathbf{r}_{\text{left hip(lh)}}, \mathbf{r}_{\text{right hip(rh)}}, \mathbf{r}_{\text{left shoulder(ls)}}, \mathbf{r}_{\text{right shoulder(rs)}}\}$. Then,

\begin{enumerate}
\item We compute the center of the predicted keypoints as $\mathbf{r}_{\text{o}} = \langle\mathbf{r}_{\text{torso}}\rangle$, where $\langle\cdot\rangle$ defines the vector average computation.
    % \begin{equation}
    %     \langle \vec{r}_{\text{o}}\rangle = \frac{1}{6}(\sum_{i}u_i, \sum_{i}v_i),
    % \end{equation}
    The resultant vector $\mathbf{r}_{\text{o}}$ is located approximately at the center of the diver's torso.

    \item We define several difference vector quantities that exist on the torso plane as
    \begin{align}
         \mathbf{r}_{\text{rsrh}} &= \mathbf{r}_{\text{rs}}-\mathbf{r}_{\text{rh}} &
          \mathbf{r}_{\text{lsrh}} &= \mathbf{r}_{\text{ls}}-\mathbf{r}_{\text{rh}} \\
          \mathbf{r}_{\text{lslh}} &= \mathbf{r}_{\text{ls}}-\mathbf{r}_{\text{lh}} & 
          \mathbf{r}_{\text{rslh}} &= \mathbf{r}_{\text{rs}}-\mathbf{r}_{\text{lh}}.
    \end{align}
    These quantities are needed to establish the relationships between joint locations, effectively defining the torso plane and conditioning the proceeding analysis with respect to the torso plane.
    
    \item We compute the diver's facing direction by taking the average direction of the cross product between the difference vectors of the torso joints. This defines a direction perpendicular to the torso plane
    \begin{align}
         \mathbf{r}_{\text{r}_{\times}} &= 
         \mathbf{r}_{\text{lsrh}} \times \mathbf{r}_{\text{rsrh}}\\
          \mathbf{r}_{\text{l}_{\times}} &= \mathbf{r}_{\text{lslh}} \times  
          \mathbf{r}_{\text{rslh}}.
    \end{align}    
    \noindent To compute the average direction and define a unit vector, we first take the average, and then we divide by the vector $L2$-norm
    \begin{equation}
        {}^{c}\hat{z}_{B} \equiv \frac{\langle\mathbf{r}_{\text{r}_{\times}},\mathbf{r}_{\text{l}_{\times}}\rangle}{\lVert\langle\mathbf{r}_{\text{r}_{\times}},\mathbf{r}_{\text{l}_{\times}}\rangle\rVert_2}.
        \label{eqn:alignment_vector}
    \end{equation}
    The alignment vector given by (\ref{eqn:alignment_vector}) points in a direction perpendicular to the plane defined by the torso keypoints. We now affix a right-handed coordinate system to $\mathbf{r}_{\text{o}}$, with ${}^{c}\hat{z}_{B}$ aligned along the direction given in (\ref{eqn:alignment_vector}). 
    % The notation means that the unit vector in the z-direction defined for the body with respect to the camera frame\starnote{check this sentence please?}. 
   We choose ${}^{c}\hat{y}_{B}$ to be the vector that points along the direction of the midpoint between hip joints. This is given by computing the midpoint of the line segment connecting the hip joints
    \begin{equation}
        \mathbf{r}_{\text{midpt}} = \langle \mathbf{r}_{\text{lh}},\mathbf{r}_{\text{rh}} \rangle.
    \end{equation}
    \item From this we compute the unit vector that points from the center of mass 
 vector $\mathbf{r}_{o}$ to $\mathbf{r}_{\text{midpt}}$. This unit vector is defined to be   ${}^{c}\hat{y}_{B}$
    \begin{equation}
         {}^{c}\hat{y}_{B} = \frac{\mathbf{r}_{\text{midpt}}-\mathbf{r}_{o}}{\lVert \mathbf{r}_{\text{midpt}}-\mathbf{r}_{o} \rVert_2}.
    \end{equation}
    \item Finally, the ${}^{c}\hat{x}_{B}$ is computed through a cross product ${}^{c}\hat{x}_{B} ={}^{c}\hat{y}_{B} \times {}^{c}\hat{z}_{B}$.
    Together these constitute the body frame ${}^{c}\mathbf{\mathcal{F}}_{B} = [{}^{c}\hat{x}_{B},{}^{c}\hat{y}_{B},{}^{c}\hat{z}_{B},\mathbf{r}_{\text{o}}] $ of the human diver, affixed to the midpoint of the extracted pose keypoints, with the ${}^{c}\hat{z}_{B}$ aligned
    in the direction perpendicular to the plane defined by the torso keypoints.
\end{enumerate}

\noindent Rotational acceleration of the body frame is an approximation to the second derivative for each of the rotational angles about the unit vectors of ${}^{c}\mathbf{\mathcal{F}}_{B}$. One can consult \citet{goldstein} for a detailed derivation of the second derivative of a rotation matrix. For brevity, we utilize the notation $\theta$ to indicate the rotation angle about the $\hat{x}$-axis, $\phi$ to indicate the rotation angle about the $\hat{y}$-axis, and $\psi$ to indicate the rotation angle about the $\hat{z}$-axis. Fig.~\ref{fig:body_frame} shows this convention.

Now let $\mathbf{X} = \{\mathbf{x}_{i}\}$, where $i=\{1,\dots,N\}$ and $N$ is the total number of time steps, define the feature vector. We then define a feature vector for a single time step as 
\begin{equation}
    \mathbf{x}_{i} = \{\ddot{x}_{i,1},\ddot{y}_{i,2},\ddot{z}_{i,3},\dots,\ddot{x}_{i,M-1},\ddot{y}_{i,M-1},\ddot{z}_{i,M-1},\ddot{\theta}_{i},\ddot{\phi}_{i},\ddot{\psi}_{i}\},
\end{equation}
where the first $1,\dots,M-1$ terms represent approximations to the second derivatives with respect to three-dimensional positions for the $M=12$ keypoints, and the last three terms are the second derivative with respect to the rotation frame of the human body. 
% \change A feature vector, including both translational and rotational features will have dimensions $N\times(3M+3)$.\stopchange
% \starnote{JS: Should have, or has? Even if we reduce it later?} 
% However, because \textit{VideoPose3D} produces three-dimensional pose estimates up to scale,
\change To remove the influence of the camera movement, \stopchange
we subtract the left hip keypoint location ($\mathbf{r}_{\text{lh}}$) from all keypoints. Therefore every keypoint is located with respect to the left hip. We then remove the left hip keypoint acceleration from the feature vector, effectively reducing the dimensionality to $N\times(3(M-1)+3)$ for both translational and rotational feature vectors. 
\jchange Along with horizontal image flipping and random rotation on 3D diver pose sequence estimates, this feature extraction pipeline is used to create a train-test dataset. \stopchange
% Using this feature extraction pipeline, along with horizontal image flipping, and random rotation on our three-dimensional pose sequence estimates, we produced $6114$ train, $1528$ validation, and $1713$ test multivariate time series samples for model training and evaluation with $N=50$ time steps. 

% We extracted feature vectors for training our temporal classification methods by tracking pose keypoint locations over a sequence of $N$ left and right stereo image pairs. Additionally we utilize finite difference approximations to estimate keypoint translational accelerations.

\begin{figure*}[h]
\vspace{1.75mm}
    \centering
   \includegraphics[width=1.0\textwidth]{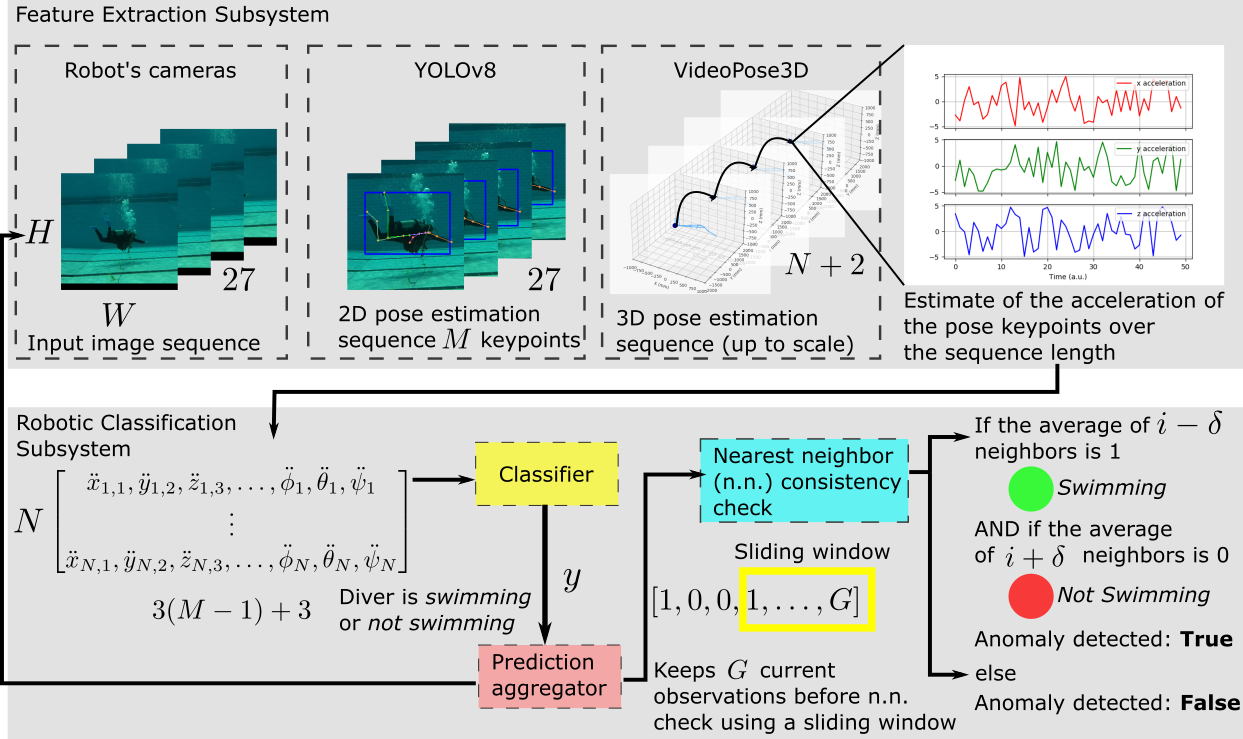}
    \caption{(Best if viewed at $2.5\times$ zoom level). Overview of the diver anomaly classification system. The system comprises two functional subsystems: a feature extractor, which creates the pseudo-IMU vector, and the robotic classifier that utilizes an arbitrary classifier for determining if the diver is \textit{swimming} or \textit{not swimming}.}
    \label{fig:methodology}
    \vspace{-4mm}
\end{figure*}
\section{Methodology}
\label{sec:methodology}
% \vspace{-1mm}
The diver anomaly classification system described in this paper relies on two subsystems: a feature extractor (described in Sec.~\ref{sec:data_collection}), which extracts pseudo-IMU values, and a classification system to perform inference based on the temporal observations of the diver's state over time. Fig.~\ref{fig:methodology} shows a summary of the methodology, where the top left block demonstrates the input image sequence, from which \textit{YOLOv8} produces two-dimensional pose estimates. \textit{VideoPose3D} requires $27$ two-dimensional pose estimates to produce a single three-dimensional estimate. \change After $N+2$ three-dimensional poses, which are required to estimate the acceleration quantities of the keypoints and body frame using \textit{central difference approximations}, we construct the feature vector. \stopchange The classifier (shown in yellow in Fig.~\ref{fig:methodology}) uses a feature vector of size $N\times (3(M-1)+3)$, $N\times 3$, or $N\times (3(M-1))$, for combined rotation and acceleration, rotation only, or translation only acceleration features, respectively, to classify the sequence as either \textit{swimming} or \textit{not swimming}. 

% \begin{figure*}[h]
%     \centering
%    \includegraphics[width=1.0\textwidth]{Figures/PNG_Sources/anomaly_detection_system_methodology.png}
%     \caption{(Best if viewed at $2.5\times$ zoom level). Overview of the diver anomaly classification system. The system comprises two functional subsystems: a feature extractor, which creates the pseudo-IMU vector, and the robotic classifier that utilizes an arbitrary classifier for determining if the diver is \textit{swimming} or \textit{not swimming}.}
%     \label{fig:methodology}
%     \vspace{-4mm}
% \end{figure*}

The prediction aggregator (shown in salmon color in Fig.~\ref{fig:methodology}) aggregates these observations. 
% Even with high testing accuracy, deep neural networks can produce erroneous results. 
To mitigate erroneous observations, we employ a nearest-neighbor consistency check (shown in blue in Fig.~\ref{fig:methodology}), which utilizes a sliding window over the latest $G$ observations. If there exists some prediction $y_j$, $j\in\{1,\dots G\}$ for which the average of the past $\delta$ neighbors is $1$ and the average of the future $\delta$ neighbors is $0$, then the diver has transitioned from \textit{swimming} to \textit{not swimming}. We empirically determined that \change $\delta=7$ \stopchange, and $G=15$ produce higher state transition accuracy. 

The primary contribution of this work is an approach to swimming state transition classification that leverages \jchange visual changes in body keypoints as a proxy for IMUs to assess diver pose.
To thoroughly evaluate the effectiveness of our pseudo-IMU features, we benchmarked them using a diverse suite of six time-series classification (TSC) models. 
This suite was chosen to represent a range of approaches, from established statistical methods to state-of-the-art deep learning architectures. \stopchange
 We utilize six different methods for TSC: time series forest based on \citet{deng2013time}, CNN and CNN Channel Wise based on \citet{zheng2014time}, CNN LSTM Dual Network from \citet{karim2019multivariate}, CNN LSTM Layer Wise from \citet{mutegeki2020cnn}, and Transformer based on \citet{zerveas2021transformer}.
We describe each method briefly below but refer the reader to the relevant literature source for more detailed explanations and network architectures.
% Table~\ref{table:classification_accuracy} shows the classification statistics for each method utilizing $6114$ train, $1528$ validation, and $1713$ test samples for training and evaluation, with $N=50$ time steps. 
% \starnote{ET: justify with citation why this is a good baseline?}

The \textbf{Time Series Forest} \jchange serves as a non-neural baseline to evaluate the inherent discriminative power of our features without complex deep learning\stopchange. It utilizes $N_{\text{trees}}=500$ trees for splitting the pseudo-IMU feature dataset nodes into the trees. A splitting criterion based on entropy gain is used to decide node splits within the tree. Additionally,~\citet{deng2013time} introduce the concept of the margin split, which helps reduce the number of candidate subsequence splits that produce the same entropy gain.

\textbf{CNN and CNN Channel Wise}~\citet{zheng2014time} introduce both a CNN-based classification architecture, which utilizes a series of stages consisting of layers of convolution, activation, and pooling layers, and a CNN Channel Wise (CNN CW) method that splits the time series into individual univariate subsequences, performs feature extraction on each subsequence, concatenates the output, and then follows with multi-layer perceptron classification. \jchange Both of the above CNN-based models were chosen to test the hypothesis that local patterns of acceleration (\eg the kick cycle) are the most important features for classification.\stopchange 

The \textbf{CNN LSTM Dual Network}~\citet{karim2019multivariate}~(CNN LSTM DN) use a CNN-based architecture for feature extraction, and then it applies an LSTM in parallel. The resulting output from both the CNN and the LSTM are concatenated and flattened for classification. We employ a notable modification from the original network architecture: unlike \citet{karim2019multivariate} who use three convolutional blocks, we achieved sufficient results with one convolutional block on both open-source datasets and underwater-specific data. 

\textbf{CNN LSTM Layer Wise}~\citet{mutegeki2020cnn} (CNN LSTM LW) architecture, in contrast to the (CNN LSTM DN) first uses convolutional layers for feature extraction and then applies an LSTM on the extracted features, rather than the raw input sequence itself. \jchange This model was included to investigate whether combining local feature extraction (CNN) with longer-term temporal dependency modeling (LSTM) would yield superior performance.\stopchange

The \textbf{Time-Series Transformer} introduced in~\citet{zerveas2021transformer}, \jchange was included to determine if the global self-attention mechanism, which can relate data points across the entire 50-step sequence, could capture complex, long-range motion dynamics that other models might miss\stopchange. %was enhanced in the following ways.
% However, a  complete treatment of the transformer architecture can be found in \cite{vaswani2017attention}. 
% was modified the transformer architecture from \cite{zerveas2021transformer} in the following ways. 
Although the original paper employs \textit{fixed} positional encodings, subsequent studies have demonstrated that \textit{learnable} positional encodings yield better results \citep{haviv2022transformer}. 
As such, \jchange we project the time series onto an embedding space using a matrix with learnable weights\stopchange~before feeding it to a self-attention block.
Additionally, while layer normalization is optimal for natural language processing, batch normalization has been shown to yield better performance when applied to numerical input \citep{yao2021leveraging}.
Therefore, within the self-attention block, we substitute batch normalization for the layer normalization layers employed by~\citet{vaswani2017attention}. 
Following \citet{zerveas2021transformer}, we begin the model training process with an unsupervised pre-training stage, where the input series is randomly masked and the model attempts to predict masked target values. 
This approach helps the network learn the shape of the data, facilitating speed-ups in the supervised learning phase. 
Pre-training is particularly useful given our constraints -- transformers demand large quantities of training data to learn representations of input information. 
Visual information is significantly more costly to collect and label than hardware-based IMU data, a bottleneck intensified by the unusually high requirements of underwater equipment. %operating . 
This limits the quantity of visual training data available, which we significantly mitigate through pre-training.
% \vspace{-1mm}
\section{Classification Method Evaluation}
\label{sec:bench_eval}
We evaluated each classification method offline using $6114$ train, $1528$ validation, and $1713$ test samples, with $N=50$ time steps, on an NVIDIA GeForce RTX 2080 Ti GPU with PyTorch $2.0.1$, CUDA $11.7$, and cuDNN $8$. 
%\footnote{Source code for all methods can be found at \url{https://github.com/IRVLab/diver_swimming_state_classification}}. 
Table~\ref{table:classification_accuracy} shows the classification accuracies comparing different features. 
We evaluated \textit{translational} only, shown in the first column of the table; \textit{rotational} only, shown in the second column; and combined \textit{rotational} and \textit{translational} features, shown in the last column of the table. 
Results revealed that \textit{translational} features produced the highest accuracies across all methods. 
The CNN channel-wise method also performed the highest for individual features. 
However, the time series forest achieved $90.83$ on the combined features, exceeding the CNN channel wise method. 
% \change We argue that \textit{translational} features yield higher accuracies, because the pose estimation has noise or ``jitter'' between successive image frames, which affects the acceleration features we extract from it. 
% We extract the rotation axis first from our definition, which depends on the joints of $\mathbf{r}_{\text{torso}}$. 
% When all four points that constitute the body frame are noisy, then the rotation axes become unstable, resulting in inaccurate rotational accelerations. 
% The \textit{translational} features take all $12$ joints into account. 
% Therefore, even if the keypoints are noisy, the aggregate effect is a mean smoothing, since translation averages over all joints. 
% This correctly results in the average behavior of when the diver is not moving, the translational acceleration is smaller compared to moving. \stopchange
% \starnote{See discussion on Overleaf comment here}
\jchange Translational features provide superior accuracy because they are inherently more robust to the ``jitter'' noise present in the pose estimation data, whereas rotational accelerations, in contrast, are highly sensitive to this noise. 
Their calculation depends on a rotation axis derived from just four torso keypoints $\mathbf{r}_{\text{torso}}$; if these points are noisy, the axis becomes unstable and corrupts the final feature. 
Conversely, our translational features are derived from the mean motion of all $12$ body keypoints. 
This averaging provides a powerful smoothing effect that filters out the noise from individual joints, resulting in a stable and reliable measure of the diver's aggregate movement. 
Consequently, when a diver is motionless, the translational acceleration correctly approaches zero, providing a much cleaner signal for classification. 
Additionally in Sec.~\ref{sec:open_water}, experimental results demonstrate how the proposed method can generalize across multiple underwater domains with the aid of a robust pose estimator (\eg\citep{wu2025stereo}).
\stopchange
\begin{table}[t]
\vspace{1mm}
\scriptsize
\renewcommand{\arraystretch}{1.1}
\caption{Comparison of classification accuracies utilizing translational, rotational, and combined acceleration features. We utilized $6114$ train, $1528$ validation, and $1713$ test samples for our model training and evaluation pipeline, with $N=50$ \change  three-dimensional pose estimates used for computing the acceleration data in the feature vector. \stopchange}
\begin{center}
\newcolumntype{R}{>{\centering\arraybackslash}X}%
\begin{tabularx}{\columnwidth}{l |R|R|R}
\toprule
    \multicolumn{4}{c}{\textbf{Bench test classification accuracy  statistics (\% correct)}$^a$}\\
    \cline{1-4}
    \textbf{Method}&\textbf{Trans. features ($50\times33$)}&\textbf{Rot. features ($50\times3$)}&\textbf{Rot.+Trans. features ($50\times36$)}\\
    \hline
    \multicolumn{1}{l|}{\textbf{Time Series Forest} \citep{deng2013time}}&90.13&71.63&\textbf{90.83}\\
    \multicolumn{1}{l|}{\textbf{CNN} \citep{zheng2014time}}&89.32&76.71&89.84\\
    \multicolumn{1}{l|}{\textbf{CNN Channel Wise} \citep{zheng2014time}}&\textbf{91.30}&\textbf{85.99}&90.37\\
    \multicolumn{1}{l|}{\textbf{CNN LSTM Dual Network} \citep{karim2019multivariate}}&82.60&52.95&82.95\\
    \multicolumn{1}{l|}{\textbf{CNN LSTM Layer Wise} \citep{mutegeki2020cnn}}&89.20&79.92&85.93\\    
    \multicolumn{1}{l|}{\textbf{Transformer}$^b$ \citep{zerveas2021transformer}}&90.83&65.38&84.18\\    
    \bottomrule
\end{tabularx}
\end{center}
\footnotesize{$^a$ All methods were trained using $50$ epochs.}\\
\footnotesize{$^b$ The transformer is pre-trained for 300 epochs using the method described in \citep{zerveas2021transformer} and fine-tuned for 50 epochs on our data.}\\
% \footnotesize{$^b$ Trained on sequence size of 12 images, or approximately one half a second of observation time.}\\
\label{table:classification_accuracy}
\vspace{-7mm}
\end{table}
% \vspace{-2mm}
\begin{figure*}[ht]
\vspace{1.75mm}
\captionsetup[subfigure]{labelformat=empty}
    \centering
    \vspace{2mm}
    \begin{subfigure}[t]{0.19\textwidth}
        \includegraphics[width=1.0\textwidth]{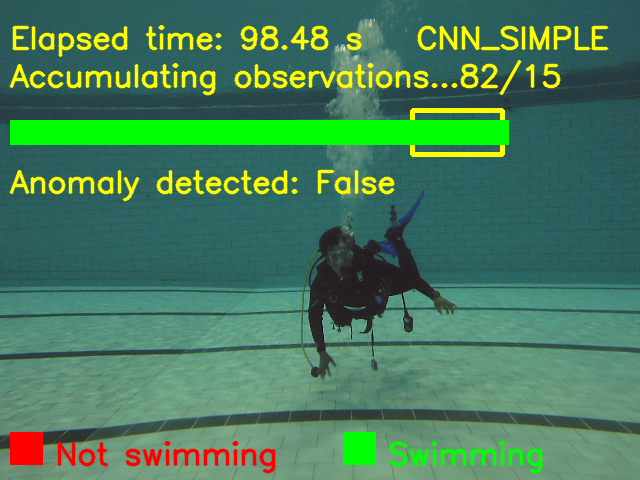}
        \caption{CNN}
    \end{subfigure}
    \begin{subfigure}[t]{0.19\textwidth}
    \includegraphics[width=1.0\textwidth]{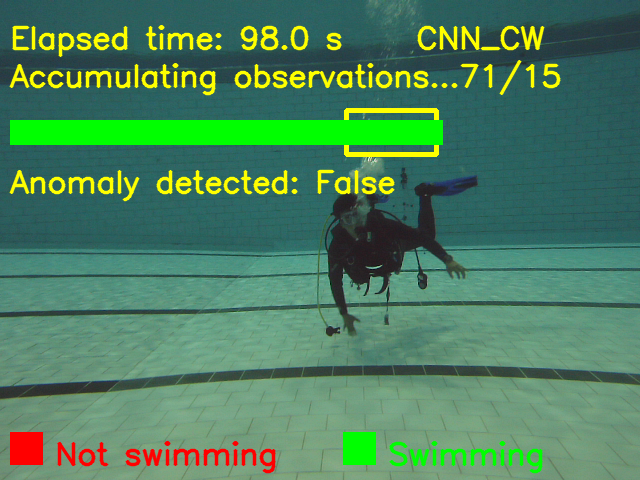}   
        \caption{CNN CW}        
    \end{subfigure}  
    \begin{subfigure}[t]{0.19\textwidth}
        \includegraphics[width=1.0\textwidth]{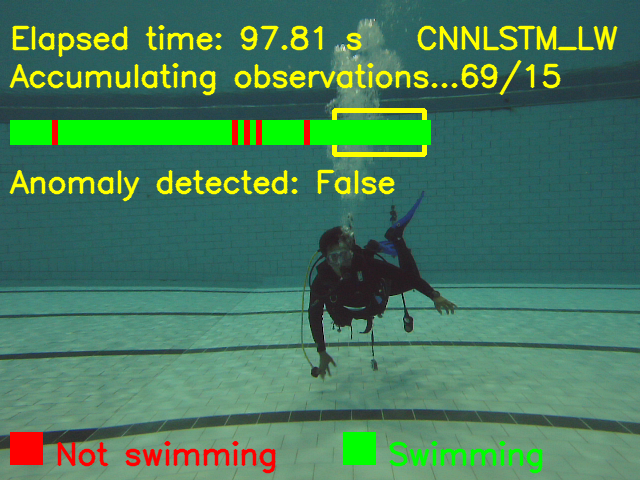}
        \caption{CNN LSTM LW}        
    \end{subfigure}
        \begin{subfigure}[t]{0.19\textwidth}
        \includegraphics[width=1.0\textwidth]{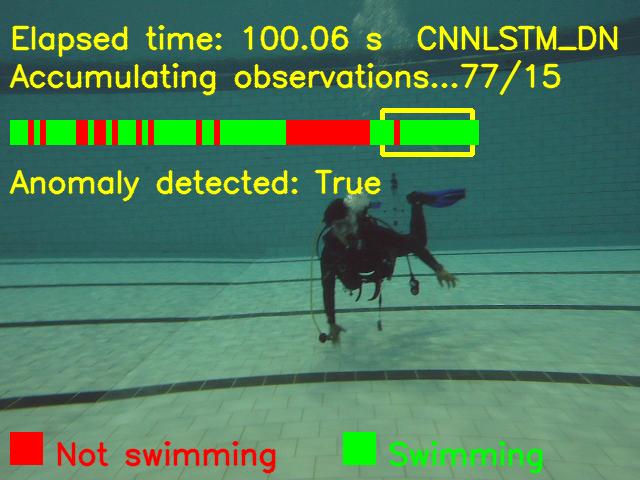}
        \caption{CNN LSTM DN}        
    \end{subfigure}
            \begin{subfigure}[t]{0.19\textwidth}
        \includegraphics[width=1.0\textwidth]{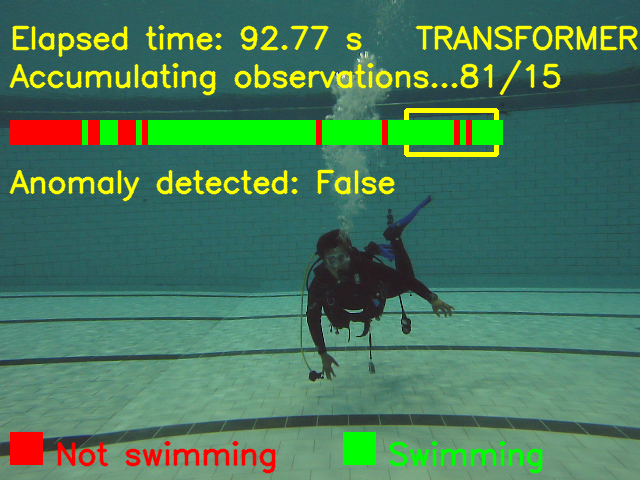}
        \caption{Transformer}        
    \end{subfigure}
    % \begin{subfigure}[t]{0.48\columnwidth}
    % \includegraphics[width=1.0\textwidth]{Figures/PNG_Sources/diver_anomaly_post_process_2024-09-13-19-45-02_CNNLSTM_DN_anomaly_50s.png}
    % \caption{CNN LSTM Dual Network}        
    % \end{subfigure}
    \caption{(Best if viewed at $4\times$ zoom). Closed-water evaluation of the swimming state classification methods.}
    \label{fig:inwater_eval}
\vspace{-5mm}
\end{figure*}
\section{\change Closed-water \stopchange Evaluation}
\label{sec:experimental_eval}
We performed in-water evaluations of the diver swimming state transition system by deploying the system onboard an AUV that uses an Nvidia Jetson TX2 embedded computing GPU. We built our codebase in a docker image \citep{merkel2014docker}, with Python3 and ROS Noetic.
% The configuration parameters are shown in the table below for the particular TX2.

We evaluated five classification methods with an experimental setup similar to the non-standard body pose data collection event. The diver swam in front of the robot's camera in a prone down position (the most common form of pose employed by divers, since it is an efficient swimming posture). The diver swam until signaled to stop. 
This ensured that the feature extractor subsystem acquired sufficient observations to produce the first feature vector \change required for classification. \stopchange
The diver then stopped all movement for $50$s. We evaluated each classifier based on its ability to classify the diver as swimming and not swimming. 
% \textbf{[Yingkun's comment: I think this can be ignored as we have already mentioned?]} 
We utilized a nearest-neighbor distance $\delta=7$ and required $G=15$ classifications before inferring if the diver experienced a transition from swimming to not swimming. 
Fig.~\ref{fig:inwater_eval} demonstrates the results taken from five of the six classification methods. 
% Results for the rest of the methods are shown in the accompanying video submission.
% \starnote{I see five of the six methods here, is this text outdated?}
Due to a software integration conflict, we were unable to deploy our time series forest classification method on the robot's TX2. 
\jchange Figures display the time elapsed from the start of the classification ROS node to the publication time of the feature vector topic\stopchange.
% In the figures, we display the elapsed time, which keeps track of the ROS time from the start of the classification ROS node to the publication time of the feature vector topic. 
Differences in elapsed time are a product of inference time for each method. 
The display shows $count/G$, which means the first $G$ \change classifications \stopchange are required before performing a nearest-neighbor check using a sliding window. \change The sliding window has a width of $\delta$ and helps us visualize the nearest-neighbor check process. \stopchange Fig.~\ref{fig:inwater_eval} shows the sliding window as a yellow bordered rectangle, which is co-linear with the observation state diagram, shown as a series of rectangles, either green or red, for \textit{swimming} or \textit{not swimming} classification, respectively. The accompanying video submission shows the sliding window moving as additional observations accumulate.   

We utilized translational features, since bench experiments revealed translational features offer the highest testing accuracy across all methods. 
% \starnote{Paraphrasing YKW: Any additional details necessary on Table 1 results?} 
Although the CNN CW performed best during bench testing, with an accuracy of $91.30\%$ (shown in Table~\ref{table:classification_accuracy}), the CNN LSTM DN performed the best during in-water evaluation, accurately determining when the diver transitioned from \textit{swimming} to \textit{not swimming}, approximately $45$~s into the evaluation experiment. Additionally, the AUV is equipped with a series of LED ring lights that communicate visual feedback of onboard processes. The AUV illuminated the lights in a green color when the diver was \textit{swimming}. Fig.~\ref{fig:cover_image} shows these lights illuminated.
\section{Open-water Field Evaluation}
\label{sec:open_water}
\begin{figure*}[ht]
\vspace{1.75mm}
\captionsetup[subfigure]{labelformat=empty}
    \centering
    \vspace{2mm}
    \begin{subfigure}[t]{0.19\textwidth}
        \includegraphics[width=1.0\textwidth]{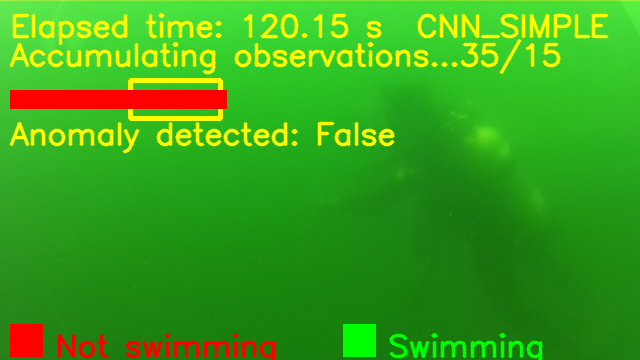}
        \caption{CNN}
    \end{subfigure}
    \begin{subfigure}[t]{0.19\textwidth}
    \includegraphics[width=1.0\textwidth]{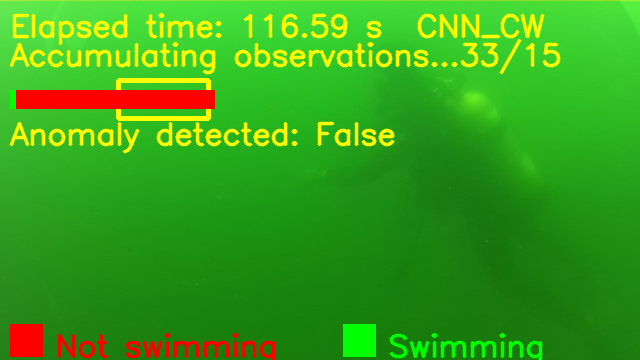}   
        \caption{CNN CW}        
    \end{subfigure}  
    \begin{subfigure}[t]{0.19\textwidth}
        \includegraphics[width=1.0\textwidth]{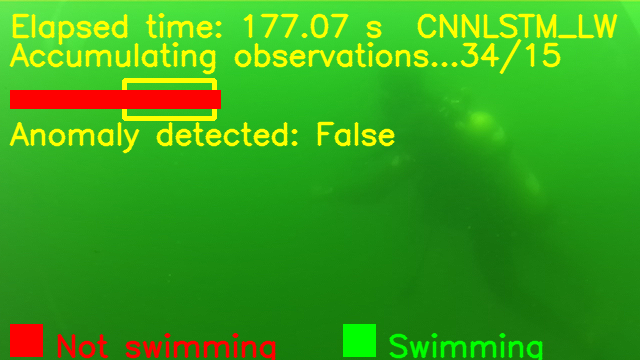}
        \caption{CNN LSTM LW}        
    \end{subfigure}
        \begin{subfigure}[t]{0.19\textwidth}
        \includegraphics[width=1.0\textwidth]{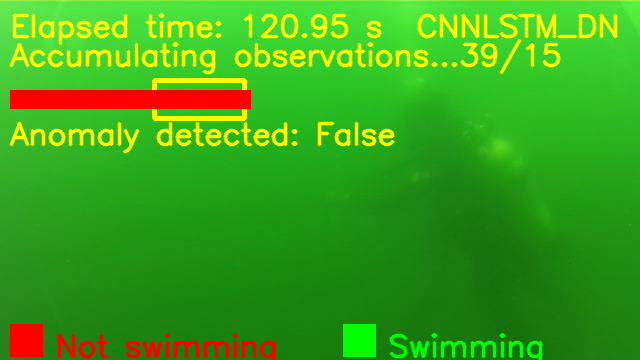}
        \caption{CNN LSTM DN}        
    \end{subfigure}
            \begin{subfigure}[t]{0.19\textwidth}
        \includegraphics[width=1.0\textwidth]{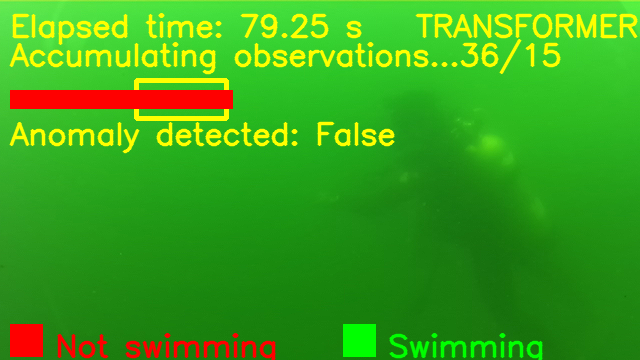}
        \caption{Transformer}        
    \end{subfigure}
    % \begin{subfigure}[t]{0.48\columnwidth}
    % \includegraphics[width=1.0\textwidth]{Figures/PNG_Sources/diver_anomaly_post_process_2024-09-13-19-45-02_CNNLSTM_DN_anomaly_50s.png}
    % \caption{CNN LSTM Dual Network}        
    % \end{subfigure}
    \caption{(Best if viewed at $4\times$ zoom). Open-water evaluation of the swimming state classification methods.}
    \label{fig:openwater_eval}
\vspace{-5mm}
\end{figure*}

We evaluated our pipeline on previously unseen data collected from a freshwater environment, in which two divers operated at a depth of five meters in approximately two meters of visibility. 
The water was turbid and green in color, which resulted in one of the worst visibility conditions. 
% The camera operator collected simulated anomalous image data using a camera with a resolution of $640\times360$ at $10$~fps. 
The camera subject was asked to perform transitions from \textit{swimming} to \textit{not swimming}. 
The \textit{swimming} portion of the data collection was $2$ minutes, and the \textit{not swimming} was $30$~s. 
The results for the five primary networks are shown in Fig.~\ref{fig:openwater_eval}. 
Notice that in all cases, the classification networks observed \textit{not swimming} states during the time period that the diver maintained a prone facing down position. 
This swimming behavior resulted in very little arm and leg movement, which could indicate why the classification methods failed to classify this as \textit{swimming} state.
% \vspace{-2mm}

\begin{figure}[ht]
\vspace{-2mm}
\captionsetup[subfigure]{labelformat=empty}
    \centering
    \vspace{2mm}
    \begin{subfigure}[t]{0.48\columnwidth}
        \includegraphics[width=1.0\textwidth]{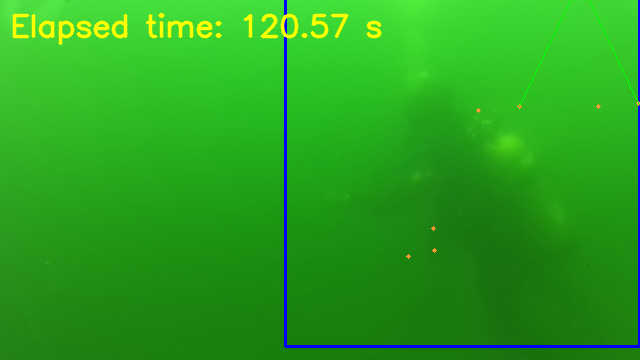}
        \caption{2D pose estimation}
    \end{subfigure}
    \begin{subfigure}[t]{0.48\columnwidth}
    \includegraphics[width=1.0\textwidth]{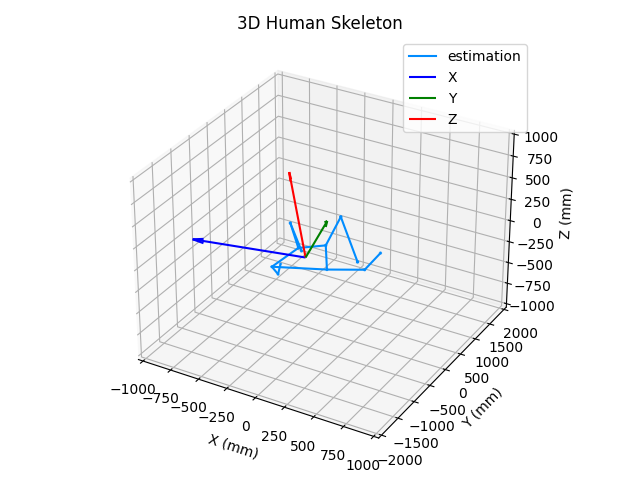}   
        \caption{3D reconstructed human pose}        
    \end{subfigure}  
    \caption{Pose estimation figures from the 2D pose estimation and resulting 3D reconstructed human pose keypoints.}
    \label{fig:openwater_2dpose}
\vspace{-6mm}
\end{figure}

These preliminary results also prove our hypothesis that the accuracy of the pose estimation network has downstream effects on our method for detecting anomalous swimming behavior. Fig.~\ref{fig:openwater_2dpose} shows that the two-dimensional estimation both fails to localize the human in the image frame and locate the human pose keypoints on the diver's torso. These results show that the pose estimation network does not transfer between environmental conditions, since the pose network was trained on closed-water swimming pool data.
\vspace{-0.5mm}

\section{Conclusion}
\label{sec:conclusion}
We introduce a novel system for robotic classification of a diver's swimming state, which is able to detect when a diver transitions from normal movement, such as kicking or paddling, to no movement. 
This mimics the state of a diver during a DI, which could lead to adverse health effects or a loss of life. 
While arrested motion is not the exclusive indicator of scuba diver distress, this work makes it possible to assess diver motion characteristics without requiring on-body sensors, while avoiding underwater data transmission challenges.
Our ongoing work is investigating a multimodal assessment approach of diver distress using additional DI indicators such as respiration rate, along with field trials in diverse regions to assess its efficacy.

% % The narrow focus of this work is a result of our assumption that DI can be modeled as a transition from swimming to not swimming, 

% since arrested motion is not the exclusive indicator of diver distress underwater. 

% \vspace{-2mm}
%\bibliographystyle{IEEEtran}
\bibliographystyle{apalike}
\bibliography{Bibliography}

\end{document}